\documentclass[sigconf]{acmart}

\renewcommand\footnotetextcopyrightpermission[1]{}
\usepackage{amsfonts}
\usepackage{algorithmic}
\usepackage{graphicx}
\usepackage{textcomp}
\usepackage{xcolor}
\usepackage{colortbl}
\usepackage{ulem}
\usepackage{url}
\usepackage{subcaption}
\usepackage{multirow}
\usepackage{balance}
\usepackage{makecell}
\usepackage{enumitem}

\begin{document}

\setlength{\columnsep}{10pt}

\newcommand{\tool}{\texttt{SeQTO}}

\title{A Selective Quantization Tuner for ONNX Models}
\titlenote{The research has been funded by Huawei Edinburgh Joint Lab Project "RobustCheck: Testing Robustness of Compiler Optimisations and Deep Learning Frameworks".}

\acmConference[ICSE '26]{The 48th International Conference on Software Engineering}{May 2026}{Rio de Janeiro, Brazil}

\author{Nikolaos Louloudakis}
\email{n.louloudakis@ed.ac.uk}
\affiliation{%
  \institution{University of Edinburgh}
  \country{United Kingdom}}

\author{Ajitha Rajan}
\email{arajan@ed.ac.uk}
\affiliation{
  \institution{University of Edinburgh}
  \country{United Kingdom}}




\begin{abstract}
Quantization reduces the precision of deep neural networks to lower model size and computational demands, but often at the expense of accuracy. Fully quantized models can suffer significant accuracy degradation, and resource-constrained hardware accelerators may not support all quantized operations. A common workaround is selective quantization, where only some layers are quantized while others remain at full precision. However, determining the optimal balance between accuracy and efficiency is a challenging task. To this direction, we propose SeQTO, a framework that enables selective quantization, deployment, and execution of ONNX models on diverse CPU and GPU devices, combined with profiling and multi-objective optimization. SeQTO generates selectively quantized models, deploys them across hardware accelerators, evaluates performance on metrics such as accuracy and size, applies Pareto Front-based objective minimization to identify optimal candidates, and provides visualization of results. We evaluated SeQTO on four ONNX models under two quantization settings across CPU and GPU devices. Our results show that SeQTO effectively identifies high-quality selectively quantized models, achieving up to $54.14$\% lower accuracy loss while maintaining up to $98.18$\% of size reduction compared to fully quantized models.
\end{abstract}

\maketitle

\vspace{-0.5em}


\section{Introduction}
Deep Neural Networks (DNNs) are widely applied in domains ranging from image recognition to content generation, offering powerful solutions to complex real-world problems. However, their large size and high computational demands pose challenges for deployment on resource-constrained hardware. Quantization~\cite{quantization} has emerged as a popular technique to reduce model size, accelerate inference, and lower power consumption by decreasing the precision of model parameters (e.g., weights and biases). Despite these benefits, quantization often leads to accuracy loss.

A common strategy to mitigate this drawback is selective quantization, where only certain layers are quantized while others remain at full precision. This approach helps balance accuracy and efficiency according to application requirements. For example, in mobile augmented reality games, developers may tolerate a small accuracy drop (e.g., $5$–$15$\%) in exchange for a substantial reduction in model size (e.g., over $70$\%), enabling deployment on lower-end devices. However, identifying this balance through tuning and profiling remains a complex task. Additionally, many hardware accelerators do not support quantization for all operations, making selective quantization not only beneficial but sometimes necessary.

To address these challenges, we propose \tool~\cite{SeQTO}\footnote{The source code of \tool, along with detailed instructions of installation and use, are available at \emph{\url{https://github.com/luludak/SeQTO}}.}, a utility for selectively quantizing ONNX~\cite{onnxsite} models with the ONNX Quantizer~\cite{onnxquantizer}, deploying them on CPU and GPU devices, and profiling performance to identify optimal candidates across multiple objectives. We validate its effectiveness through experiments on four classification models, including low-end devices such as Mali GPUs. \tool\ is designed to help developers and researchers achieve performance gains from quantization while maintaining accuracy within acceptable bounds, even under hardware-imposed constraints. The primary aim of our work is towards addressing inefficient model quantization. Our methodology aims to address inefficient quantization by reducing the search space of poorly quantized layers, iteratively excluding such layers from the optimized model while evaluating their impact, and providing profiling capabilities through multi-objective minimization, as well as visual presentation of results.

\vspace{-5pt}
\section{Related Work}
Extensive research has examined DNN quantization, with surveys reviewing advances in reducing model size and resource use while preserving accuracy~\cite{Gholami2021ASO, weng2023neuralnetworkquantizationefficient, SurveyQuantization, liu2025lowbitmodelquantizationdeep, nagel2020downadaptiveroundingposttraining}. Many works address quantization during training~\cite{esser2020learnedstepsizequantization, LowBitQuantization, nagel2022overcomingoscillationsquantizationawaretraining, Jacob_2018_CVPR, huang2022sdqstochasticdifferentiablequantization, yang2020fracbitsmixedprecisionquantization, dong2019hawqhessianawarequantization, dong2019hawqv2hessianawaretraceweighted, przewlockarus2022poweroftwoquantizationlowbitwidth, wu2018mixedprecisionquantizationconvnets}. However, the training process is costly and resource intensive, and applying quantization might require model retraining to adapt in the new values associated with it. For that matter, we also solely focus on post-training quantization.

For selective post-training quantization in particular, works such as Banner et al.~\cite{PTQ4Bit} on 4-bit post-training, Krishnamoorthi~\cite{krishnamoorthi2018quantizingdeepconvolutionalnetworks} on per-channel 8-bit quantization present the benefits of utilizing quantization down to specific bits. However, these works assume uniform post-training quantization, which can cause significant accuracy loss depending on the model structure and parameters. To address this, various mixed-precision methods have been proposed. Schaefer et al.~\cite{schaefer2023mixedprecisionposttraining} proposes a hardware-agnostic methodology for appropriate bit-width selection using performance metrics (quantization error, performance degradation and layer metadata).
Diao et al.~\cite{diao2022attentionroundposttrainingquantization} propose a mixed-precision method that applies probabilistic analysis and modifies model weights accordingly, while Hubara et al.~\cite{hubara2020improvingposttrainingneural} optimize bit-widths by utilizing integer programming. Pandley et al.~\cite{mixedprecisionqualcomm} applies an exploratory approach to detect which precision type for each layer is more suitable to minimize accuracy loss, while Ma et al.~\cite{OMPQ} considers model graph dependencies and structure to apply mixed precision. Finally, QNNRepair~\cite{QNNRepair} applies constraint solving combined with unit testing to detect and "repair" specific neuron values that degrade accuracy in quantized TFLite models~\cite{tensorflow} in the post-training phase.

Prior work has not practically addressed selective quantization strategies to balance accuracy and model size according to user requirements, a central focus of our methodology. Pandley et al.~\cite{mixedprecisionqualcomm}, the closest to ours, is limited to static quantization, relies on AIMET~\cite{siddegowda2022neuralnetworkquantizationai} rather than targeting hardware accelerators directly, lacks ONNX support, and is not publicly available. Their evaluation uses Bit-Operations (BOPs) as a proxy for efficiency rather than real-device measurements, limiting comparability. QNNRepair~\cite{QNNRepair} similarly aims to preserve accuracy and performance but follows a best-effort approach, without leveraging acceptable accuracy–performance trade-offs as we do, and is also incompatible with ONNX. 
Finally, Schaefer et al.~\cite{schaefer2023mixedprecisionposttraining} inspired our use of metrics for layer ranking to reduce the search space; however, we extend this idea by applying it to real-life hardware accelerators, a direction not explored in their work.

To the best of our knowledge, \tool\ is the first methodology to offer a comprehensive feature set for practical selective quantization tuning, including: (1) multi-objective optimization via Pareto Front analysis (using accuracy and model size objectives), (2) support for both static and dynamic quantization (Section~\ref{subsec:sel-quant}), (3) deployment on CPUs (ONNX Runtime~\cite{onnxruntime}) and GPUs (Apache TVM~\cite{tvm}), while supporting resource-constrained devices (4) full ONNX compatibility across diverse models and architectures, (5) layer analysis and ranking using multiple error metrics to guide selective quantization, and (6) visualization of the process to help users identify the quantization candidate that best meets their objective requirements.
We focus on the ONNX standard because it is widely used, supports both static and dynamic quantization, and represents models in a generic, well-defined format. This enables our approach to be directly applicable to real-world models while remaining largely model independent.

\vspace{-5pt}
\section{Methodology}
\label{s:methodology}

\begin{figure*}[!htp]
\centering
\includegraphics[width=0.85\textwidth]{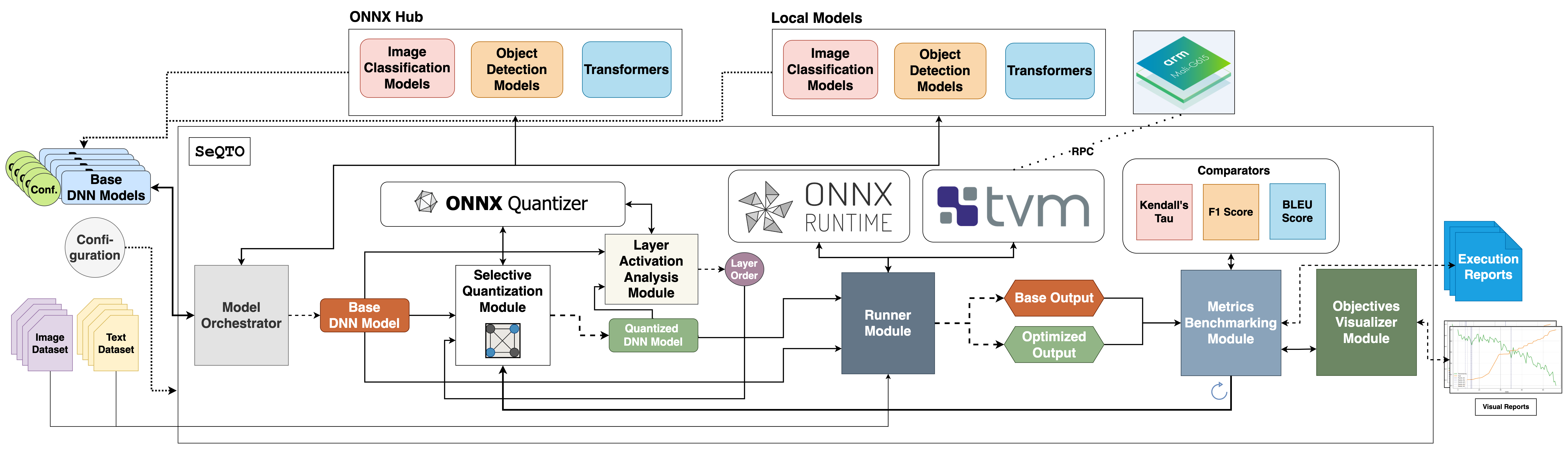}
  \vspace{-10pt}
  \caption{Architecture of \tool\ with the following primary modules: (1) Model Orchestrator, (2) Selective Quantization Module, (3) Runner Module, (4) Metrics Benchmarking Module, and (5) Objectives Visualizer Module.}
  \label{fig:SeQTO}
  \vspace{-12pt}
\end{figure*}

The architecture of \tool\ is shown in Figure~\ref{fig:SeQTO}. Its workflow loads a model, analyzes layers to identify error-prone ones, applies quantization, and executes both original and quantized models on the dataset for comparison. This process is repeated while selectively excluding layers from quantization, generating multiple candidates. The results are compared to identify top-K Pareto-optimal solutions~\cite{pareto, NDS}, minimizing objectives such as accuracy loss and model size. Finally, results are presented through visual and textual reports. \tool\ consists of five core modules: (1) Model Orchestration, (2) Selective Quantization, (3) Model Deployment \& Execution, (4) Metrics Benchmarking, and (5) Objectives Visualization, detailed in the following sections.

\vspace{-5pt}
\subsection{Model Orchestration}
\label{orch-module}

Model orchestration involves fetching the models under test along with their configurations, quantizing them wherever necessary and preparing them for deployment and  execution. \tool\ supports loading ONNX models either locally or from online repositories. In the latter case, \tool\ uses the official ONNX Model Hub~\cite{onnxmodelhub} via its API, enabling support for other third-party ONNX repositories. It also allows fetching and loading multiple models simultaneously. Loaded models and configurations are then passed to the Optimizer and Runner Modules.

\vspace{-8pt}
\subsection{Selective Quantization}
\label{subsec:sel-quant}
To enable selective quantization, \tool\ leverages the ONNX Quantizer~\cite{onnxquantizer}, which takes a model and a list of layers to exclude, producing a selectively quantized model. The ONNX Quantizer supports two methods: \textit{Static} and \textit{Dynamic Quantization}. Both quantize weights and biases at compile time; Static Quantization also quantizes activations using a calibration dataset, while Dynamic Quantization keeps activations in floating point and quantizes them on-the-fly at runtime. Using this mechanism, \tool\ generates models where only selected layers are quantized, allowing the pipeline to evaluate how quantizing or skipping specific layers affects objective minimization. Users can also explicitly exclude layers through configuration.

\vspace{-5pt}
\subsection{Layer Activation Analysis}
\label{subsec:act-analysis}
\begin{figure}[!htp]
\centering
\includegraphics[width=0.88\linewidth, height=4.2cm]{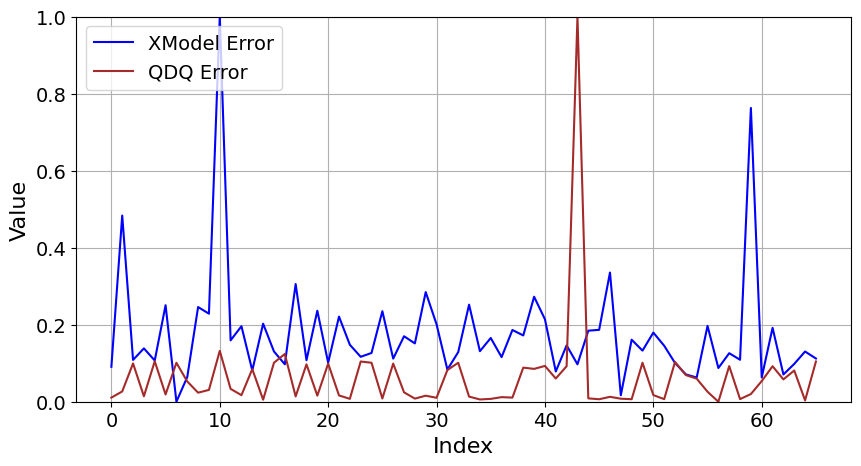}
\vspace{-15pt}
 \caption{Normalized \texttt{XModel Error} and \texttt{QDQ Error} across layers for original and quantized MobileNetV2. X-axis indicates layer index; Y-axis indicates normalized error value.}
 \label{fig:MobileNetV2Errors}
 \vspace{-15pt}
\end{figure}

Effective selective quantization requires analyzing layer-wise sensitivity to quantization error. \tool\ reduces the search space by fully quantizing the model and using a small, configurable dataset to compare activations and identify the most affected layers. Layer matching is then performed via the ONNX Quantizer API to determine which layers are most prone to quantization errors. It calculates two key metrics: \textit{QDQ Error} (\texttt{qdq\_err}) and \textit{XModel Error} (\texttt{xmodel\_err}). QDQ Error estimates how much the output of each layer changes when quantized in isolation, using a calibration dataset to predict potential accuracy loss at the layer level. XModel Error measures the contribution of each layer to the overall accuracy loss of the fully quantized model, computed as $(|output\_fp32 - output\_xmodel| / |output\_fp32|)$. Figure~\ref{fig:MobileNetV2Errors} illustrates that the predicted sensitivity of a layer may differ from its actual impact on the overall accuracy loss of the model. Consequently, we use both metrics when ranking layers for selective quantization.
We calculate an error metric for each layer by normalizing each error value between $0$ and $1$ and calculating $error\_metric = 0.5 * norm\_xmodel\_err + 0.5 * norm\_qdq\_err$. Using this metric, the layers are ranked in descending order and are utilized for the selective quantization process, essentially considering layer sensitivity and accuracy degradation. The idea is straightforward: Layer activation analysis mirrors deductive reasoning by ranking layers according to error suspiciousness, measured by the degree of problematic behavior introduced by quantization as reflected in error metrics.

\vspace{-5pt}
\subsection{Model Deployment \& Execution}
Each selectively quantized model is run on a dataset in order to detect crashes caused by operation incompatibilities with the target device. In addition, the results are  compared with the original model to minimize objectives such as accuracy loss and size. The Runner Module handles deployment and execution on the selected devices, supporting (1) ONNX Runtime~\cite{onnxruntime} for CPUs and (2) Apache TVM~\cite{tvm} for GPUs, where TVM compiles models to shared libraries and uses RPC, including on Android devices. Finally, if a crash occurs upon model execution, it is logged, and the layer activation analysis (Section~\ref{subsec:act-analysis}) results are utilized in an attempt to identify the layers likely responsible, which are then excluded from quantization, in order to enable selective quantization on resource-constrained hardware accelerators.
Currently, \tool\ performs inference on local image recognition datasets, with plans to support datasets from packages like \texttt{datasets}~\cite{datasets} for other domains (e.g., text generation, sentiment analysis). The Runner Module also supports state saving across modules for seamless resumption of interrupted runs, a key feature for low-end devices (e.g., Mali GPUs) where RPC may be unreliable. Run details, including initial layer error estimates, excluded layers, and metrics (e.g., model size and accuracy), are recorded in a metadata file. Finally, large input batches can be split into smaller chunks, with chunk size configurable via system settings.

\vspace{-5pt}
\subsection{Model Benchmarking}
\label{sub:benchmarking}
After deploying and executing both models, \tool\ uses a comparator to compute accuracy differences and extract quantized model information such as size. It currently supports top-K classification and can be extended to other model types (e.g., object detectors, transformers). Results are stored in a \texttt{JSON} report and updated after each selective quantization iteration. It then computes the top-K Pareto-optimal solutions using \texttt{NonDominatedSorting} (NDS)~\cite{NDS} from the Pymoo library~\cite{pymoo}, minimizing a number of objectives. In our setup ($K = 3$), the top solution represents the Pareto-optimal trade-off across both objectives, while the other two correspond to the best solutions for each individual objective (accuracy loss, model size). The \texttt{JSON} report serves as a checkpoint and can be directly integrated into automated pipelines (e.g., CI/CD) or used to generate visual reports, as discussed in Section~\ref{sub:visualization}.

\vspace{-8pt}
\subsection{Objectives Visualization}
\label{sub:visualization}
The main purpose of \tool\ is to provide clear recommendations for the best candidate that minimizes the desired objectives, primarily through visual reports. To this end, \tool\ supports two main operations: (1) layer-wise visualizations of error metrics for a specified model (e.g., Figure~\ref{fig:MobileNetV2Errors}), and (2) visualization of each objective obtained from the Model Benchmarking report in a unified plot, with the top three Pareto-optimal solutions (Section~\ref{sub:benchmarking}) indicated as vertical lines. An example of this operation is shown for ResNet50 in Figure~\ref{fig:comparisons}. Each visual report provides a complete representation of the tuning process, enabling users to identify the best model candidate.

\vspace{-5pt}
\section{Preliminary Results}

\begin{figure}[t]
    \centering
    \begin{subfigure}{0.45\textwidth}
        \centering
        \includegraphics[width=\textwidth, height=4.5cm]{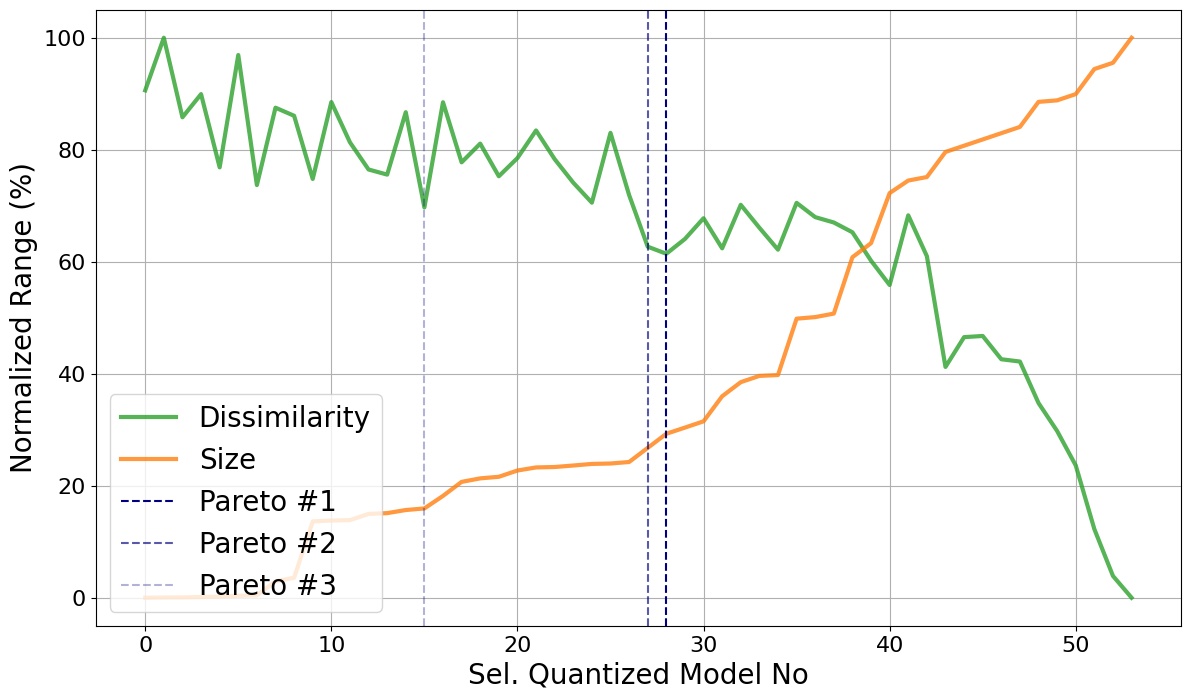}
        \caption{CPU - Dynamic Quantization}
        \label{app:resnet0sovu}
        \end{subfigure}%
  \hspace{0.05\textwidth}
    \begin{subfigure}{0.45\textwidth}
        \centering
        \includegraphics[width=\textwidth, height=4.5cm]{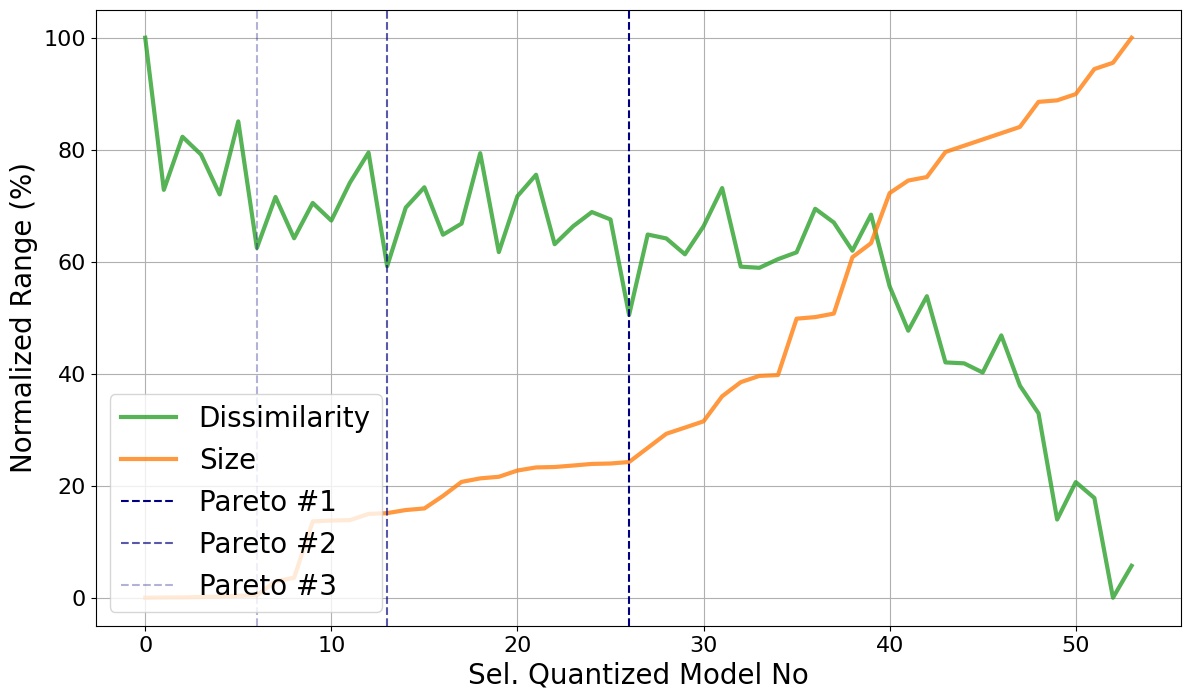}
        \caption{GPU - Dynamic Quantization}
        \label{app:resnet0local}
    \end{subfigure}%
  \label{app:resnetdnns0}
  \vspace{-10pt}
 \caption{Selective quantization of ResNet50 across hardware: X-axis shows excluded layers; Y-axis shows normalized accuracy loss (model dissimilarity) and model size.}
  \label{fig:comparisons}
  \vspace{-15pt}
\end{figure}

\setlength\extrarowheight{1pt}
\setlength{\tabcolsep}{5pt} 

\setlength{\tabcolsep}{1.5pt}
\begin{table*}[!htp]
    \centering
    \vspace{-5pt}
    \caption{Comparison of the absolute difference percentage (Ab. D. \%) across fully (FQ.) and selectively quantized models on the top Pareto Front using Static (P. St.) and Dynamic (P. Dyn.) Quantization, deployed on CPU and GPU across all tested models.}
    \label{accuracy-diff}
    \vspace{-10pt}
    \begin{tabular}{|c|c|c|c|c|c|c|c|c|c|c|c|c|}
    \hline
        \textbf{Model} & \multicolumn{6}{|c|}{\textbf{CPU}} & \multicolumn{6}{|c|}{\textbf{GPU}} \\ \hline
        ~ & \textbf{FQ. St.} & \textbf{P. St.} & \textbf{Ab. D. \%} & \textbf{FQ. Dyn.} & \textbf{P. Dyn.} & \textbf{Ab. D. \%} & \textbf{FQ. St.} & \textbf{P. St.} & \textbf{Ab. D. \%} & \textbf{FQ. Dyn.} & \textbf{P. Dyn.} & \textbf{Ab. D. \%} \\ \hline
        \textbf{MobileNetV2} & 45.28\% & 34.1\% & \cellcolor{orange!20}24.69\% & 40.8\% & 23.12\% & \cellcolor{orange!75}43.33\% & 36.65\% & 26.19\%  & \cellcolor{orange!20}28.54\% & 35.23\% & 18.09\% & \cellcolor{orange!80}48.65\% \\ \hline
        \textbf{ShuffleNetV2} & 43.45\% & 26.77\% & \cellcolor{orange!70}38.39\% & 32.44\% & 15.87\% & \cellcolor{orange!90}51.08\% & 35.53\% & 19.55\% & \cellcolor{orange!75}44.98\% & 26.58\% & 12.19\%  & \cellcolor{orange!100}54.14\% \\ \hline
        \textbf{ENet-Lite4} & 54.11\% & 27.38\% & \cellcolor{orange!80}49.40\% & 43.26\% & 21.41\% & \cellcolor{orange!100}50.51\% & 43.71\% & 26.45\% & \cellcolor{orange!70}39.49\% & 27.52\% & 21.09\% & \cellcolor{orange!20}23.36\% \\ \hline
        \textbf{ResNet50} & 37.2\% & 27.86\%  & \cellcolor{orange!20}25.11\% & 25.91\% & 19.08\% & \cellcolor{orange!20}26.36\% & 28.94\% & 17.69\% & \cellcolor{orange!70}38.87\% & 21.46\% & 12.07\% & \cellcolor{orange!75}43.76\% \\ \hline
    \end{tabular}
\end{table*}

\setlength{\tabcolsep}{1.5pt}
\begin{table*}[!ht]
    \centering
    \vspace{-5pt}
    \caption{Comparison of the model size (in MB) across fully (FQ.) and selectively (partially (P.)) quantized models on the top Pareto Front using Static (St.) and Dynamic (D.) Quantization, based on CPU/GPU back-ends across all tested models. \texttt{Pres \%} presents how much of the quantized model size reduction was achieved (``preserved"), compared to the fully quantized model.}
    \vspace{-10pt}
    \label{model-size-diff}
    \begin{tabular}{|c|c|c|c|c|c|c|c|c|c|c|c|}
    \hline
        \textbf{Model} & \multicolumn{3}{|c|}{\textbf{Init./FQ.}} & \multicolumn{4}{|c|}{\textbf{CPU}} & \multicolumn{4}{|c|}{\textbf{GPU}} \\ \hline
        ~ & \textbf{Init.} & \textbf{FQ. St.} & \textbf{FQ. Dyn.} & \textbf{P. St.} & \textbf{Pres. \%} & \textbf{P. Dyn.} & \textbf{Pres. \%} & \textbf{P. St.} & \textbf{Pres. \%} & \textbf{P. Dyn.} & \textbf{Pres. \%} \\ \hline
        \textbf{MobileNetV2} & 13.32 MB & 3.43 MB & 3.43 MB & 7.34 MB & \cellcolor{orange!60}60.47\% & 3.75 MB & \cellcolor{orange!100}96.76\% & 5.67 MB & \cellcolor{orange!75}77.35\% & 3.61 MB & \cellcolor{orange!100}98.18\% \\ \hline
        \textbf{ShuffleNetV2} & 8.8 MB & 2.4 MB & 2.36 MB & 4.7 MB & \cellcolor{orange!60}64.06\% & 4.78 MB & \cellcolor{orange!60}62.42\% & 4.73 MB & \cellcolor{orange!60}63.59\% & 4.83 MB & \cellcolor{orange!60}61.65\% \\ \hline
        \textbf{ENet-Lite4} & 49.54 MB & 12.86 MB & 12.87 MB & 33.69 MB & \cellcolor{orange!43}43.21\% & 36.13 MB & \cellcolor{orange!35}36.57\% & 20.94 MB & \cellcolor{orange!75}77.97\% & 19.62 MB & \cellcolor{orange!80}81.59\% \\ \hline
        \textbf{ResNet50} & 97.82 MB & 24.96 MB & 24.92 MB & 33.2 MB & \cellcolor{orange!85}88.69\% & 44.55 MB & \cellcolor{orange!70}73.07\% & 41.31 MB & \cellcolor{orange!75}77.56\% & 41.17 MB & \cellcolor{orange!75}77.71\% \\ \hline
    \end{tabular}
    \vspace{-8pt}
\end{table*}

\begin{description}[leftmargin=0pt]
    \item[Experiment Set:]{
    Using \tool, we conducted experiments on four well-known classification models with varying ONNX opset versions (op.): \texttt{MobileNetV2}~\cite{mobilenetv2} (op.=10), \texttt{ShuffleNetV2}~\cite{shufflenet} (op.=10), \texttt{EfficientNet-Lite4} (ENet-Lite4)~\cite{efficientnet} (op.=11), and \texttt{ResNet50}~\cite{resnet} (op.=12). We sourced the models from the official ONNX Model Hub~\cite{onnxmodelhub}. The models were compiled and deployed on two hardware acceleration devices: an \texttt{Intel-i5 CPU} (in an HP laptop) and a \texttt{Mali-G68 MC4 GPU} (in a Xiaomi Redmi Note 11 Pro+ mobile device). We refer to them as \texttt{CPU} and \texttt{GPU} hereafter. All models were tested using both Static and Dynamic quantization, as described in section~\ref{s:methodology}. For evaluation, we used the \texttt{ILSVRC2017}~\cite{ILSVRC17} validation dataset containing 5500 images. To maintain tractability, each iteration randomly selected 300 images for accuracy evaluation, and up to 50 images for Static Quantization calibration. Our optimization objectives were model accuracy loss and model size. Accuracy loss is measured as the dissimilarity between the original and quantized top predictions of a model, reflecting precision loss in the model weights and the resulting degradation in top-class prediction accuracy.
    }
    \item [Visual Reporting:] {
    \tool\ was able to determine the Pareto Front for each of the models and generate the visual reports for each system. As an indicative example, we present the visual reports for \texttt{ResNet50} on both CPU and GPU using Dynamic Quantization, as generated by \tool\ in Figure~\ref{fig:comparisons}\footnote{The full set of plots from our experiments is available in the \tool\ code repository under the ``results" folder.}. 
    The $X$ axis demonstrates the number of excluded layers, ranging from fully quantized model (no excluded layers) towards excluding all model layers.The $Y$ axis demonstrates the normalized percentage for each objective, applied so that different objectives can be included in the same figure. Finally, the top $3$ Pareto Fronts are presented in the form of vertical lines. \tool\ records excluded layers in the $JSON$ report, allowing users to generate the corresponding model variant for each instance presented in the visual report.
    }
    \item [Objective Minimization:] {
    We also present the top Pareto Front candidate selection along with its results for the objectives aforementioned. Table~\ref{accuracy-diff} presents the accuracy percentage differences for the top label prediction, and Table~\ref{model-size-diff} contains differences in model size (in MB). In particular, each table reports the absolute percentage differences between fully and selectively quantized models under both static and dynamic quantization, with results shown for CPU and GPU execution.
    Overall, \tool\ successfully identified candidates that significantly reduced accuracy loss of up to $54.14$\% (for ShuffleNetV2 on GPU using Dynamic Quantization), compared to the fully quantized model. It also identified candidates that preserved most of the model size reductions  achieved by quantization (up to $98.18$\%, for MobileNetV2 on GPU using Dynamic Quantization). Overall, we observed that \tool\ selected different model choices for its Pareto-optimal objectives. While model size reduction remained consistent across CPU and GPU backends, the impact of quantization on accuracy varied by hardware.
    }
\end{description}

\vspace{-10pt}
\section{Conclusions and Future Work}
We presented \tool, a utility for tuning selective quantization of ONNX models, and demonstrated its effectiveness on four models deployed on CPU and GPU. \tool\ reduced accuracy loss by up to $54.14$\% while retaining up to $98.18$\% of the model size reduction achieved by full quantization.

Future work will extend \tool\ to additional architectures, including object detectors (e.g., Tiny YOLOv3~\cite{YOLOv3}, SSD~\cite{SSD}) and transformers (e.g., T5~\cite{T5Paper}, GPT-2~\cite{gpt2}), and support evaluation metrics such as F1~\cite{F1Score} and BLEU~\cite{BLEU}. We also plan to add explicit support for hardware bit-width constraints (e.g., INT8, INT4, mixed precision). Our goal is to help users select ONNX model candidates that best balance accuracy, size, and performance under hardware constraints. Given the widespread adoption of ONNX, we believe \tool\ can significantly advance DNN quantization and deployment. Finally, we note the importance of additional quantization metrics, such as time overhead. Although \tool\ can already measure these metrics and apply Pareto-front minimization, their systematic evaluation is left for future work.

\balance

\bibliographystyle{ACM-Reference-Format}
\bibliography{00-main.bib}

@misc{onnxsite,
  title={{Open Neural Network Exchange}},
  year={2023},
  howpublished={\url{https://onnx.ai}},
  note={[Accessed 23-Apr-2025]}
}

@misc{pymoo,
  title={{pymoo: Multi-objective Optimization in Python}},
  year={2025},
  howpublished={\url{https://pymoo.org/}},
  note={[Accessed 8-Jul-2025]}
}

@article{pareto,
author = {Das, Indraneel and Dennis, J. E.},
title = {{Normal-Boundary Intersection: A New Method for Generating the Pareto Surface in Nonlinear Multicriteria Optimization Problems}},
journal = {SIAM Journal on Optimization},
volume = {8},
number = {3},
pages = {631-657},
year = {1998},
doi = {10.1137/S1052623496307510},
URL = {https://doi.org/10.1137/S1052623496307510},
eprint = {https://doi.org/10.1137/S1052623496307510}
}

@InProceedings{QNNRepair,
author="Song, Xidan
and Sun, Youcheng
and Mustafa, Mustafa A.
and Cordeiro, Lucas C.",
editor="Ferreira, Carla
and Willemse, Tim A. C.",
title="QNNRepair: Quantized Neural Network Repair",
booktitle="Software Engineering and Formal Methods",
year="2023",
publisher="Springer Nature Switzerland",
address="Cham",
pages="320--339",
isbn="978-3-031-47115-5"
}

@misc{mixedprecisionqualcomm,
      title={{A Practical Mixed Precision Algorithm for Post-Training Quantization}}, 
      author={Nilesh Prasad Pandey and Markus Nagel and Mart van Baalen and Yin Huang and Chirag Patel and Tijmen Blankevoort},
      year={2023},
      eprint={2302.05397},
      archivePrefix={arXiv},
      primaryClass={cs.LG},
      url={https://arxiv.org/abs/2302.05397}, 
}

@inproceedings{quantization,
  author       = {Song Han and
                  Huizi Mao and
                  William J. Dally},
  editor       = {Yoshua Bengio and
                  Yann LeCun},
  title        = {{Deep Compression: Compressing Deep Neural Network with Pruning, Trained
                  Quantization and Huffman Coding}},
  booktitle    = {4th International Conference on Learning Representations, {ICLR} 2016,
                  San Juan, Puerto Rico, May 2-4, 2016, Conference Track Proceedings},
  year         = {2016},
  url          = {http://arxiv.org/abs/1510.00149},
  bibsource    = {dblp computer science bibliography, https://dblp.org}
}

@misc{nagel2020downadaptiveroundingposttraining,
      title={{Up or Down? Adaptive Rounding for Post-Training Quantization}}, 
      author={Markus Nagel and Rana Ali Amjad and Mart van Baalen and Christos Louizos and Tijmen Blankevoort},
      year={2020},
      eprint={2004.10568},
      archivePrefix={arXiv},
      primaryClass={cs.LG},
      url={https://arxiv.org/abs/2004.10568}, 
}

@misc{onnxquantizer,
  title={{ONNX Quantizer}},
  year={2019},
  howpublished={\url{https://onnxruntime.ai/docs/performance/model-optimizations/quantization.html}},
  note={[Accessed 23-Apr-2025]}
}

@misc{onnxmodelhub,
  title={{ONNX Model Hub}},
  year={2023},
  howpublished={\url{https://github.com/onnx/onnx/blob/main/docs/Hub.md}},
  note={[Accessed 23-Apr-2025]}
}

@misc{onnxruntime,
  title={{ONNX Runtime}},
  year={2019},
  howpublished={\url{https://onnxruntime.ai}},
  note={[Accessed 23-Apr-2025]}
}

@inproceedings{tvm,
  title = {{{{TVM}}: {{An}} Automated End-to-End Optimizing Compiler for Deep Learning}},
  booktitle = {13th {{USENIX}} Symposium on Operating Systems Design and Implementation ({{OSDI}} 18)},
  author = {Chen, Tianqi and Moreau, Thierry and Jiang, Ziheng and Zheng, Lianmin and Yan, Eddie and Shen, Haichen and Cowan, Meghan and Wang, Leyuan and Hu, Yuwei and Ceze, Luis and Guestrin, Carlos and Krishnamurthy, Arvind},
  year = {2018},
  month = oct,
  pages = {578--594},
  isbn = {978-1-939133-08-3}
}

@misc{datasets,
  title={{Datasets}},
  year={2020},
  howpublished={\url{https://huggingface.co/docs/datasets}},
  note={[Accessed 23-Apr-2025]}
}

@article{mobilenetv2,
  author    = {Mark Sandler and
               Andrew G. Howard and
               Menglong Zhu and
               Andrey Zhmoginov and
               Liang{-}Chieh Chen},
  title     = {{Inverted Residuals and Linear Bottlenecks: Mobile Networks for Classification, Detection and Segmentation}},
  journal   = {CoRR},
  volume    = {abs/1801.04381},
  year      = {2018},
  url       = {http://arxiv.org/abs/1801.04381},
  eprinttype = {arXiv},
  eprint    = {1801.04381},
  bibsource = {dblp computer science bibliography, https://dblp.org}
}

@inproceedings{shufflenet,
author = {Zhang, Xiangyu and others},
year = {2018},
month = {06},
pages = {6848-6856},
title = {{ShuffleNet: An Extremely Efficient Convolutional Neural Network for Mobile Devices}},
booktitle = {2018 IEEE/CVF CVPR},
doi = {10.1109/CVPR.2018.00716}
}

@misc{efficientnet,
      title={{EfficientNet: Rethinking Model Scaling for Convolutional Neural Networks}}, 
      author={Mingxing Tan and others},
      year={2020},
      eprint={1905.11946},
      archivePrefix={arXiv},
      primaryClass={cs.LG}
}

@article{resnet,
  author    = {Kaiming He and
               Xiangyu Zhang and
               Shaoqing Ren and
               Jian Sun},
  title     = {{Deep Residual Learning for Image Recognition}},
  journal   = {CoRR},
  volume    = {abs/1512.03385},
  year      = {2015},
  url       = {http://arxiv.org/abs/1512.03385},
  eprinttype = {arXiv},
  eprint    = {1512.03385},
  bibsource = {dblp computer science bibliography, https://dblp.org}
}

@article{ILSVRC17,
Author = {Olga Russakovsky and Jia Deng and Hao Su and Jonathan Krause and Sanjeev Satheesh and Sean Ma and Zhiheng Huang and Andrej Karpathy and Aditya Khosla and Michael Bernstein and Alexander C. Berg and Li Fei-Fei},
Title = {{ImageNet Large Scale Visual Recognition Challenge}},
Year = {2015},
journal   = {International Journal of Computer Vision (IJCV)},
doi = {10.1007/s11263-015-0816-y},
volume={115},
number={3},
pages={211-252}
}

@article{T5Paper,
author = {Raffel, Colin and Shazeer, Noam and Roberts, Adam and Lee, Katherine and Narang, Sharan and Matena, Michael and Zhou, Yanqi and Li, Wei and Liu, Peter J.},
title = {Exploring the limits of transfer learning with a unified text-to-text transformer},
year = {2020},
issue_date = {January 2020},
publisher = {JMLR.org},
volume = {21},
number = {1},
issn = {1532-4435},
journal = {J. Mach. Learn. Res.},
month = jan,
articleno = {140},
numpages = {67}
}

@inproceedings{gpt2,
  title={{Language Models are Unsupervised Multitask Learners},
  author={Alec Radford and Jeff Wu and Rewon Child and David Luan and Dario Amodei and Ilya Sutskever}},
  year={2019},
  url={https://api.semanticscholar.org/CorpusID:160025533}
}

@inproceedings{BLEU,
author = {Papineni, Kishore and Roukos, Salim and Ward, Todd and Zhu, Wei-Jing},
title = {{BLEU: a method for automatic evaluation of machine translation}},
year = {2002},
publisher = {Association for Computational Linguistics},
address = {USA},
url = {https://doi.org/10.3115/1073083.1073135},
doi = {10.3115/1073083.1073135},
booktitle = {Proceedings of the 40th Annual Meeting on Association for Computational Linguistics},
pages = {311–318},
numpages = {8},
location = {Philadelphia, Pennsylvania},
series = {ACL '02}
}

@article{F1Score,
author = {Christen, Peter and Hand, David J. and Kirielle, Nishadi},
title = {A Review of the F-Measure: Its History, Properties, Criticism, and Alternatives},
year = {2023},
issue_date = {March 2024},
publisher = {Association for Computing Machinery},
address = {New York, NY, USA},
volume = {56},
number = {3},
issn = {0360-0300},
url = {https://doi.org/10.1145/3606367},
doi = {10.1145/3606367},
journal = {ACM Comput. Surv.},
month = oct,
articleno = {73},
numpages = {24}
}

@article{Gholami2021ASO,
  title={A Survey of Quantization Methods for Efficient Neural Network Inference},
  author={Amir Gholami and Sehoon Kim and Zhen Dong and Zhewei Yao and Michael W. Mahoney and Kurt Keutzer},
  journal={ArXiv},
  year={2021},
  volume={abs/2103.13630},
  url={https://api.semanticscholar.org/CorpusID:232352683}
}

@misc{weng2023neuralnetworkquantizationefficient,
      title={Neural Network Quantization for Efficient Inference: A Survey}, 
      author={Olivia Weng},
      year={2023},
      eprint={2112.06126},
      archivePrefix={arXiv},
      primaryClass={cs.LG},
      url={https://arxiv.org/abs/2112.06126}, 
}

@article{SurveyQuantization,
author = {Rokh, Babak and Azarpeyvand, Ali and Khanteymoori, Alireza},
title = {A Comprehensive Survey on Model Quantization for Deep Neural Networks in Image Classification},
year = {2023},
issue_date = {December 2023},
publisher = {Association for Computing Machinery},
address = {New York, NY, USA},
volume = {14},
number = {6},
issn = {2157-6904},
url = {https://doi.org/10.1145/3623402},
doi = {10.1145/3623402},
journal = {ACM Trans. Intell. Syst. Technol.},
month = nov,
articleno = {97},
numpages = {50}
}

@misc{liu2025lowbitmodelquantizationdeep,
      title={Low-bit Model Quantization for Deep Neural Networks: A Survey}, 
      author={Kai Liu and Qian Zheng and Kaiwen Tao and Zhiteng Li and Haotong Qin and Wenbo Li and Yong Guo and Xianglong Liu and Linghe Kong and Guihai Chen and Yulun Zhang and Xiaokang Yang},
      year={2025},
      eprint={2505.05530},
      archivePrefix={arXiv},
      primaryClass={cs.LG},
      url={https://arxiv.org/abs/2505.05530}, 
}

@article{YOLOv3,
  title={{YOLOv3: An Incremental Improvement}},
  author={Joseph Redmon and Ali Farhadi},
  journal={ArXiv},
  year={2018},
  volume={abs/1804.02767},
  url={https://api.semanticscholar.org/CorpusID:4714433}
}

@InProceedings{SSD,
author="Liu, Wei
and Anguelov, Dragomir
and Erhan, Dumitru
and Szegedy, Christian
and Reed, Scott
and Fu, Cheng-Yang
and Berg, Alexander C.",
editor="Leibe, Bastian
and Matas, Jiri
and Sebe, Nicu
and Welling, Max",
title="SSD: Single Shot MultiBox Detector",
booktitle="Computer Vision -- ECCV 2016",
year="2016",
publisher="Springer International Publishing",
address="Cham",
pages="21--37",
isbn="978-3-319-46448-0"
}

@inproceedings{tensorflow,
author = {Abadi, Mart\'{\i}n and Barham, Paul and Chen, Jianmin and Chen, Zhifeng and Davis, Andy and Dean, Jeffrey and Devin, Matthieu and Ghemawat, Sanjay and Irving, Geoffrey and Isard, Michael and Kudlur, Manjunath and Levenberg, Josh and Monga, Rajat and Moore, Sherry and others},
title = {TensorFlow: a system for large-scale machine learning},
year = {2016},
isbn = {9781931971331},
publisher = {USENIX Association},
address = {USA},
booktitle = {Proceedings of the 12th USENIX Conference on Operating Systems Design and Implementation},
pages = {265–283},
numpages = {19},
location = {Savannah, GA, USA},
series = {OSDI'16}
}

@inbook{PTQ4Bit,
author = {Banner, Ron and Nahshan, Yury and Soudry, Daniel},
title = {Post training 4-bit quantization of convolutional networks for rapid-deployment},
year = {2019},
publisher = {Curran Associates Inc.},
address = {Red Hook, NY, USA},
booktitle = {Proceedings of the 33rd International Conference on Neural Information Processing Systems},
articleno = {714},
numpages = {9}
}

@misc{krishnamoorthi2018quantizingdeepconvolutionalnetworks,
      title={Quantizing deep convolutional networks for efficient inference: A whitepaper}, 
      author={Raghuraman Krishnamoorthi},
      year={2018},
      eprint={1806.08342},
      archivePrefix={arXiv},
      primaryClass={cs.LG},
      url={https://arxiv.org/abs/1806.08342}, 
}

@misc{esser2020learnedstepsizequantization,
      title={{Learned Step Size Quantization}}, 
      author={Steven K. Esser and Jeffrey L. McKinstry and Deepika Bablani and Rathinakumar Appuswamy and Dharmendra S. Modha},
      year={2020},
      eprint={1902.08153},
      archivePrefix={arXiv},
      primaryClass={cs.LG},
      url={https://arxiv.org/abs/1902.08153}, 
}

@INPROCEEDINGS{LowBitQuantization,
  author={Bhalgat, Yash and Lee, Jinwon and Nagel, Markus and Blankevoort, Tijmen and Kwak, Nojun},
  booktitle={2020 IEEE/CVF Conference on Computer Vision and Pattern Recognition Workshops (CVPRW)}, 
  title={LSQ+: Improving low-bit quantization through learnable offsets and better initialization}, 
  year={2020},
  volume={},
  number={},
  pages={2978-2985},
  doi={10.1109/CVPRW50498.2020.00356}}

@misc{nagel2022overcomingoscillationsquantizationawaretraining,
      title={Overcoming Oscillations in Quantization-Aware Training}, 
      author={Markus Nagel and Marios Fournarakis and Yelysei Bondarenko and Tijmen Blankevoort},
      year={2022},
      eprint={2203.11086},
      archivePrefix={arXiv},
      primaryClass={cs.LG},
      url={https://arxiv.org/abs/2203.11086}, 
}

@InProceedings{Jacob_2018_CVPR,
author = {Jacob, Benoit and Kligys, Skirmantas and Chen, Bo and Zhu, Menglong and Tang, Matthew and Howard, Andrew and Adam, Hartwig and Kalenichenko, Dmitry},
title = {Quantization and Training of Neural Networks for Efficient Integer-Arithmetic-Only Inference},
booktitle = {Proceedings of the IEEE Conference on Computer Vision and Pattern Recognition (CVPR)},
month = {June},
year = {2018}
}

@misc{OMPQ,
      title={{OMPQ: Orthogonal Mixed Precision Quantization}}, 
      author={Yuexiao Ma and Taisong Jin and Xiawu Zheng and Yan Wang and Huixia Li and Yongjian Wu and Guannan Jiang and Wei Zhang and Rongrong Ji},
      year={2022},
      eprint={2109.07865},
      archivePrefix={arXiv},
      primaryClass={cs.LG},
      url={https://arxiv.org/abs/2109.07865}, 
}

@misc{huang2022sdqstochasticdifferentiablequantization,
      title={SDQ: Stochastic Differentiable Quantization with Mixed Precision}, 
      author={Xijie Huang and Zhiqiang Shen and Shichao Li and Zechun Liu and Xianghong Hu and Jeffry Wicaksana and Eric Xing and Kwang-Ting Cheng},
      year={2022},
      eprint={2206.04459},
      archivePrefix={arXiv},
      primaryClass={cs.LG},
      url={https://arxiv.org/abs/2206.04459}, 
}

@misc{yang2020fracbitsmixedprecisionquantization,
      title={{FracBits: Mixed Precision Quantization via Fractional Bit-Widths}}, 
      author={Linjie Yang and Qing Jin},
      year={2020},
      eprint={2007.02017},
      archivePrefix={arXiv},
      primaryClass={cs.CV},
      url={https://arxiv.org/abs/2007.02017}, 
}

@misc{dong2019hawqhessianawarequantization,
      title={HAWQ: Hessian AWare Quantization of Neural Networks with Mixed-Precision}, 
      author={Zhen Dong and Zhewei Yao and Amir Gholami and Michael Mahoney and Kurt Keutzer},
      year={2019},
      eprint={1905.03696},
      archivePrefix={arXiv},
      primaryClass={cs.CV},
      url={https://arxiv.org/abs/1905.03696}, 
}

@misc{dong2019hawqv2hessianawaretraceweighted,
      title={HAWQ-V2: Hessian Aware trace-Weighted Quantization of Neural Networks}, 
      author={Zhen Dong and Zhewei Yao and Yaohui Cai and Daiyaan Arfeen and Amir Gholami and Michael W. Mahoney and Kurt Keutzer},
      year={2019},
      eprint={1911.03852},
      archivePrefix={arXiv},
      primaryClass={cs.CV},
      url={https://arxiv.org/abs/1911.03852}, 
}

@misc{wu2018mixedprecisionquantizationconvnets,
      title={{Mixed Precision Quantization of ConvNets via Differentiable Neural Architecture Search}}, 
      author={Bichen Wu and Yanghan Wang and Peizhao Zhang and Yuandong Tian and Peter Vajda and Kurt Keutzer},
      year={2018},
      eprint={1812.00090},
      archivePrefix={arXiv},
      primaryClass={cs.CV},
      url={https://arxiv.org/abs/1812.00090}, 
}

@misc{przewlockarus2022poweroftwoquantizationlowbitwidth,
      title={{Power-of-Two Quantization for Low Bitwidth and Hardware Compliant Neural Networks}}, 
      author={Dominika Przewlocka-Rus and Syed Shakib Sarwar and H. Ekin Sumbul and Yuecheng Li and Barbara De Salvo},
      year={2022},
      eprint={2203.05025},
      archivePrefix={arXiv},
      primaryClass={cs.LG},
      url={https://arxiv.org/abs/2203.05025}, 
}

@misc{schaefer2023mixedprecisionposttraining,
      title={{Mixed Precision Post Training Quantization of Neural Networks with Sensitivity Guided Search}}, 
      author={Clemens JS Schaefer and Elfie Guo and Caitlin Stanton and Xiaofan Zhang and Tom Jablin and Navid Lambert-Shirzad and Jian Li and Chiachen Chou and Siddharth Joshi and Yu Emma Wang},
      year={2023},
      eprint={2302.01382},
      archivePrefix={arXiv},
      primaryClass={cs.LG},
      url={https://arxiv.org/abs/2302.01382}, 
}

@misc{diao2022attentionroundposttrainingquantization,
      title={Attention Round for Post-Training Quantization}, 
      author={Huabin Diao and Gongyan Li and Shaoyun Xu and Yuexing Hao},
      year={2022},
      eprint={2207.03088},
      archivePrefix={arXiv},
      primaryClass={cs.LG},
      url={https://arxiv.org/abs/2207.03088}, 
}

@InProceedings{NDS,
author="Deb, Kalyanmoy
and Agrawal, Samir
and Pratap, Amrit
and Meyarivan, T.",
editor="Schoenauer, Marc
and Deb, Kalyanmoy
and Rudolph, G{\"u}nther
and Yao, Xin
and Lutton, Evelyne
and Merelo, Juan Julian
and Schwefel, Hans-Paul",
title="A Fast Elitist Non-dominated Sorting Genetic Algorithm for Multi-objective Optimization: NSGA-II",
booktitle="Parallel Problem Solving from Nature PPSN VI",
year="2000",
publisher="Springer Berlin Heidelberg",
address="Berlin, Heidelberg",
pages="849--858",
isbn="978-3-540-45356-7"
}

@misc{hubara2020improvingposttrainingneural,
      title={Improving Post Training Neural Quantization: Layer-wise Calibration and Integer Programming}, 
      author={Itay Hubara and Yury Nahshan and Yair Hanani and Ron Banner and Daniel Soudry},
      year={2020},
      eprint={2006.10518},
      archivePrefix={arXiv},
      primaryClass={cs.LG},
      url={https://arxiv.org/abs/2006.10518}, 
}

@misc{siddegowda2022neuralnetworkquantizationai,
      title={Neural Network Quantization with AI Model Efficiency Toolkit (AIMET)}, 
      author={Sangeetha Siddegowda and Marios Fournarakis and Markus Nagel and Tijmen Blankevoort and Chirag Patel and Abhijit Khobare},
      year={2022},
      eprint={2201.08442},
      archivePrefix={arXiv},
      primaryClass={cs.LG},
      url={https://arxiv.org/abs/2201.08442}, 
}

@misc{SeQTO,
  title        = {SeQTO},
  author       = {Nikolaos Louloudakis},
  year         = {2025},
  publisher    = {Zenodo},
  doi          = {10.5281/zenodo.18155147},
  url          = {https://zenodo.org/records/18155147}
}

\end{document}